\documentclass[10pt,twocolumn,letterpaper]{article}

\usepackage[pagenumbers]{wacv} 

\definecolor{wacvblue}{rgb}{0.21,0.49,0.74}
\usepackage[pagebackref,breaklinks,colorlinks,allcolors=wacvblue]{hyperref}
\usepackage{algorithm}
\usepackage{algpseudocode}
\usepackage{placeins}
\usepackage{float} 

\def\wacvPaperID{3077} 
\def\confName{WACV}
\def\confYear{2027}

\title{Sufficient Frame Subsets for Video Understanding}

\author{Lahiru Wijayasingha\\
Independent Researcher\\
{\tt\small xlahirunuwan@gmail.com}\\
{\tt\small \url{https://cyclopesvision.com}}
}

\begin{document}
\maketitle
\begin{abstract}
Understanding when during a video a model makes its critical observations is essential for interpreting video-understanding systems. However, existing attribution methods often spread evidence diffusely across near-duplicate frames. We introduce Sufficient Frame Subsets: compact sets of frames in which all other frames are masked, yet each subset independently preserves the model’s prediction. We evaluate our method on two complementary benchmarks, UCF101 and Something-Something v2, using both CNN- and transformer-based architectures. Our results show that the proposed algorithm identifies highly compact and effective frame subsets for explaining model decisions. We also find that models capable of relying on a greater number of sufficient frame subsets tend to perform better, possibly because they are more robust, consistent with prior research. Finally, the extracted frame subsets help us analyze model failure cases, reveal which aspects of a video the model has learned to emphasize, and identify opportunities for improving future training procedures.
\end{abstract}
    
\section{Introduction}
\label{sec:intro}

Beyond spatial attention, a fundamental question in video-understanding AI is when, during a video, a model makes its critical observations. Answering this question is essential for interpreting and gaining professional acceptance of these systems, particularly in clinical settings  \cite{alceu_bissoto_fd5ccb3c, hyunkyung_han_75679c56}. We show that multiple disjoint subsets of frames may each independently be sufficient for the model to reach the same prediction. Building on the sufficient input subset framework introduced by Carter et al.  \cite{brandon_carter_b02b21dd}, we introduce \textbf{Sufficient Frame Subsets }\textbf{(SFSs)} for video understanding. Multiple SFSs can reveal the diverse reasoning pathways available to the model, offering a novel lens for interpretability and a diagnostic tool for assessing model robustness.

Conventional attribution techniques such as Grad-CAM  \cite{ramprasaath_r__selvaraju_509d91f2}, Integrated Gradients  \cite{mukund_sundararajan_62086538} and Shapley values  \cite{anja_stadlhofer_9b5ed5fc}  were originally used in computer vision primarily for static images, where they localize the input regions that most influence a model’s prediction. These methods have since been extended to the temporal domain, typically by applying them frame by frame and aggregating the resulting heatmaps. However, such direct adaptations are of limited value for video understanding. Treating each frame as an independent image neglects the temporal interplay among frames, which is crucial for video understanding such as for action recognition  \cite{avinab_saha_e5427afc}, and can produce unsatisfactory explanations  \cite{zhenqiang_li_45594fd6}. This challenge is further compounded by the strong redundancy of consecutive frames in many videos  \cite{haixin_huang_6600015c}. Because neighboring frames are highly correlated, attributions can spread diffusely across near-duplicate frames  \cite{zerui_wang_6183c62e} hindering the interpretability. Our method avoids this limitation by reformulating explanation as a selection problem rather than a pixel-level attribution problem. Instead of assigning an importance value to every frame, it distills the model’s decision into several sufficient subsets each containing only a few frames, each of which can independently reproduce the model’s output  \cite{brandon_carter_b02b21dd}. The resulting explanations are compact, temporally meaningful, and directly interpretable.

\begin{figure*}[t]
	\includegraphics[width=\textwidth]{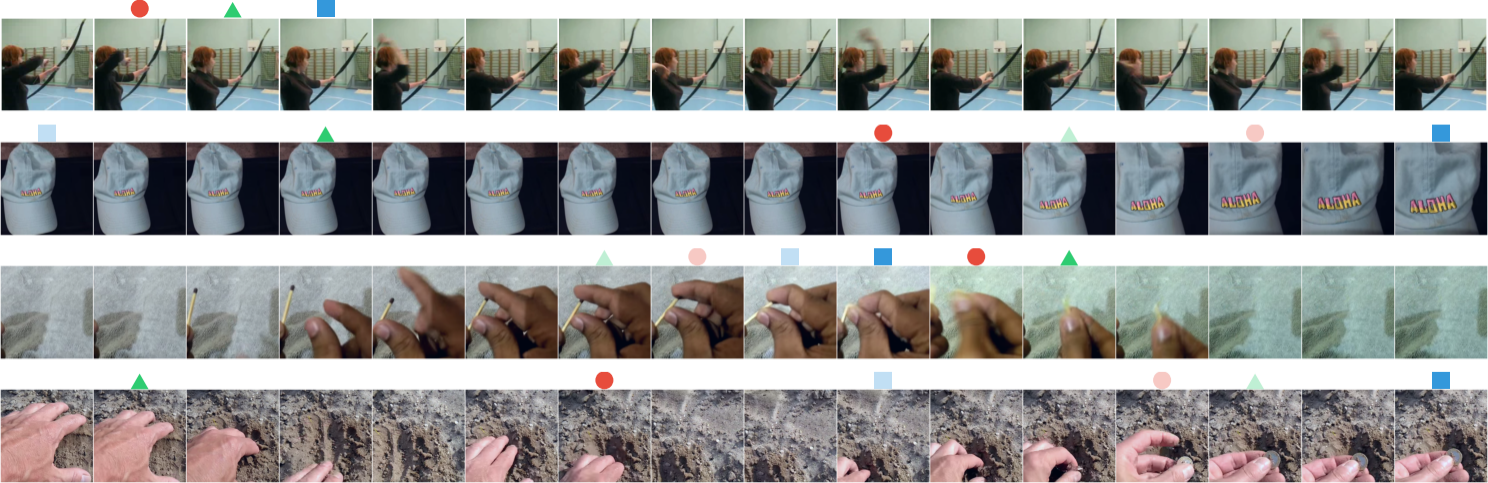}
	\caption{Several Examples for detected SFS: from top to bottom: Archery (UCF101),  Approaching something with your camera, Bending something until it breaks, Digging something out of something (SSV2). Different shapes on the frames indicate different SFSs extracted from the video. We limit the number of SFS shown to three to maintain clarity.}
	\label{fig:example}
\end{figure*}

At the heart of our approach lies the sufficient input subset framework of Carter et al.  \cite{brandon_carter_b02b21dd}. A sufficient input subset is an approximately minimal subset of input features whose observed values alone suffice for a black-box model to produce the same decision, even if no other information about the remaining input is provided. To identify such subsets, Carter et al. propose an entirely model-agnostic, instance-wise backward selection procedure that iteratively masks the least-important remaining feature until the smallest subset satisfying a sufficiency criterion is found. To the best of our knowledge, we are the first to introduce this framework to the video domain, extracting sufficient frame subsets directly from video inputs. Concretely, we search over a binary selection mask for an approximately minimal set of frames whose retained values alone reproduce the model's decision. A frame subset is then deemed sufficient when the relative drop in the predicted-class probability stays below a threshold, mirroring the sufficiency criterion introduced by Carter et al.  \cite{brandon_carter_b02b21dd}.

Play Fair addresses a closely related problem by attributing a video model’s decision to its constituent frames  \cite{will_price_ec92d5a0}. Our objective is substantially different in both its explanatory target and its methodological assumptions. Play Fair identifies an ordering of frames according to their attributed importance, whereas our method searches for one or more sufficient frame subsets: compact sets of frames whose retained evidence enables the model to reach the same decision as it reaches when given the complete video. This distinction is important because videos contain substantial temporal redundancy. Ranking individual frames can therefore produce noisy or unstable attributions when several neighboring frames convey overlapping evidence. By evaluating subsets through decision preservation, our method identifies combinations of frames that are jointly sufficient rather than merely assigning an importance score to each frame. 

Our method uses the established technique of occlusion sensitivity analysis to estimate the importance of groups of frames. This approach occludes portions of the input using a neutral baseline value and measures the resulting change in the model’s output. A substantial decrease in the classification probability suggests that the occluded features are important for the prediction  \cite{matthew_d__zeiler_d7ae7e07, mengnan_du_9dc59bf6, melkamu_abay_mersha_3aaed7eb}. Occlusion sensitivity analysis have been applied in the video domain mainly for spacial attribution \cite{tomoki_uchiyama_b5106143}.

Applying occlusion sensitivity analysis to a set of video frames requires evaluating the model after those frames have been removed or otherwise excluded from the input  \cite{will_price_ec92d5a0}. This presents a practical challenge because most video-classification models are designed to process inputs with a fixed temporal length and cannot directly handle shortened videos  \cite{elahe_vahdani_b8cb360c}. A method for evaluating variable-length inputs is therefore necessary to determine whether removing particular frames changes the model’s prediction. Play Fair addresses this issue by constructing a multi-scale model that accommodates videos of varying temporal lengths; consequently, its explanations describe the decisions of that modified model  \cite{will_price_ec92d5a0}. Rather than altering the model to accommodate shortened videos, our method preserves the original input length and temporal structure by replacing masked frames through temporal forward-fill. Specifically, it repeats the closest preceding retained frame instead of using a fixed baseline or synthetic value. This approach adapts the sufficient-input-subset principle  \cite{brandon_carter_b02b21dd} to temporally coherent video data while limiting the distributional shift that can distort explanations  \cite{ruth_c__fong_7e902699}.

Our contributions are threefold. First, we extend the sufficient input subset framework to the video domain by introducing sufficient \emph{frame} subsets. Specifically, we demonstrate that multiple frame subsets can be extracted from a video and that, when presented to the model, each subset independently preserves the model’s original decision. Second, we show that our algorithm can identify collections of higher quality sufficient frame subsets more efficiently than commonly used frame-attribution methods. Finally, we demonstrate that these extracted subsets can help reveal and analyze diverse failure modes in video-understanding models.

\section{Related Work}
\label{sec:formatting}

An adjacent line of work explains video models through removal or perturbation rather than gradient propagation, yet none of it is framed in terms of decision preservation. Gaya-Morey et al. proposed REVEX, a unified framework that re-expresses removal-based explanation techniques for video within a common formalization, covering occlusion, perturbation, and subset-selection styles of analysis  \cite{f__xavier_gaya_morey_82e6f4cd}. Li et al. extended perturbation-based attribution to video by sampling spatio-temporal masks and imposing temporal-smoothness constraints, so that attributions respect the continuity of motion rather than flickering across near-duplicate frames  \cite{zhenqiang_li_7ef0b643}. In a compressed-domain setting, Srinivasan et al. applied layer-wise relevance propagation to action recognition and observed that, in many cases, only a small number of significant frames drive the model's decision  \cite{vignesh_srinivasan_3519fcb1}. Within the gradient family, Grad-CAM++ generalized the Grad-CAM weighting scheme for more faithful visual explanations  \cite{aditya_chattopadhay_629bcc87}. These methods measure influence or importance; none of them asks which subset of frames \textit{alone suffices} to reproduce the prediction, the question that defines our frame subsets.

The problem of selecting a small subset of video content that preserves as much predictive information as possible is naturally studied through submodular maximization, which formalizes the diminishing-returns property shared by coverage-like objectives  \cite{andreas_krause_89c76cc5} \cite{uriel_feige_8e8fa037}. Facility-location is a subset of submodular maximization which is applicable for selecting meaningful input tokens for video classification model \cite{janghoon_cho_62c5bc88}. For scalability, sampling-based variants of these algorithms exist \cite{baharan_mirzasoleiman_d4b72f29}.
\section{Methods}

A video can be represented as an ordered sequence of frames, $V = \{I_t\}_{t=1}^{T}$\textit{. Given a video-understanding model }$\Phi$\textit{, let }$\Phi(V)_c$\textit{ denote the corresponding output probability for class }$c$\textit{. Let }$\mathcal{S} \subseteq V$\textit{ be a subset of frames selected from the video. We define a sufficient frame subset as a set of frames }$\mathcal{S}$\textit{ that is sufficient to preserve the output probability for a given class. Let the output probability obtained from the video with all frames present be }$\Phi(V)_c$\textit{. When only the subset of frames }$\mathcal{S}$\textit{ is retained, we denote the resulting video as }$V_{\mathcal{S}}$. Note that there may be several Sufficient Frame Subsets (SFS) like $\mathcal{S}$, which we call the Sufficient Frame Subset Collection, SFS-COL = $\{S_1, \ldots, S_K\}$. Our algorithm can find SFS-COL, as described in Section \ref{algo}. Each SFS in SFS-COL i.e $S_k$ for $k=\{1,…,K\}$ must satisfy the following sufficiency condition :

\begin{equation}
	\frac{\Phi(V)_c - \Phi(V_{\mathcal{S}})_c}{\Phi(V)_c} \leq \tau
\end{equation}

This is the analogue of the SIS sufficiency criterion  \cite{brandon_carter_b02b21dd} except that we calculate the relative decrease in probability where as Carter et.al. examine the value of the output probability $\Phi(V_{\mathcal{S}})_c \geq \tau^{/} $. We made this change because we wanted a more standardized metric that is transferable among different models, datasets and applications. 

video-understanding models are sensitive to the order of the frames: the same content at different temporal positions can yield different decisions. We ask which temporal evidence a model actually relies on. A subset of frames $S$ is a \textbf{sufficient frame subset} if, when every frame outside $S$ is absent, the model's confidence for the predicted class remains above a threshold $\tau$ the retained frames alone reproduce the decision. This follows the sufficiency–minimality paradigm established for arbitrary feature sets  \cite{brandon_carter_b02b21dd} and spatial regions  \cite{keisuke_kawano_0114a8f7}; our contribution is its temporal instantiation, where frames are ordered, mutually redundant, and motion-coupled. We additionally require that each SFS be minimal (no proper subset is itself sufficient) and that the detected SFS collection be complete: no set of frames outside it preserves the decision under any masking. Exact minimality is intractable, so we relax it as in prior work   \cite{brandon_carter_b02b21dd},  \cite{keisuke_kawano_0114a8f7} and describe in Section \ref{sec:ordering_qaulity} the frame-sorting algorithm and imputation baselines used to obtain sufficient frame subsets.

\subsection{Sufficient frame subset collection algorithm}
\label{algo}
We obtain sufficient frame subsets using an instance-wise backward selection strategy adapted from the sufficient input subset framework of Carter et al. \cite{brandon_carter_b02b21dd}. We begin with the full set of frames and iteratively identify the frame whose removal results in the highest model confidence for the given class, masking it out and repeating until every frame has been eliminated; the frames are thereby ranked from least to most decisive. We then rebuild a sufficient frame subset by re-adding frames in reverse order of removal. The frames dropped last are the ones whose presence mattered most, until the relative drop in model's confidence once again reaches $\tau$. Because each backward-selection query evaluates $\Phi$ over the joint set of remaining frames, the procedure accounts for interactions between frames, an important consideration in video, where temporally adjacent frames are near-duplicates and it maintains sufficiency of the remaining set throughout its execution, unlike a forward selection that greedily accumulates marginal gains. Masked frames are replaced by the neutral baseline frame produced by our frame imputation method. Having found one frame subset, we mask out all of its frames and repeat the entire procedure on the remaining frames, collecting further disjoint frame subsets until the frames outside the collection no longer support the decision on their own. As in the original formulation, this yields a completeness property for the collection, while worst-case cost is $O(p^2 k)$ model evaluations for $k$ frame subsets, reducible to $O(Tk)$ by batching the argmax queries on a GPU where $T$ is the number of frames in the video.

\subsection{Frame imputation methods}

Sufficient input subsets were originally introduced and evaluated in text, image, and genomic data, while video was explicitly outside the scope of the framework  \cite{brandon_carter_b02b21dd}. In these domains, replacing masked values is comparatively forgiving: the original method substitutes removed features with a baseline, such as the feature mean, zeros, or hot-deck samples drawn from the feature distribution, and reports that these alternatives generally produce similar predictions in practice  \cite{brandon_carter_b02b21dd}. This relative robustness reflects the fact that static images and bag-of-words representations lack the temporal structure that masking can disrupt. Video, however, is fundamentally different. Common interpretation baselines including constant grey patches  \cite{matthew_d__zeiler_d7ae7e07}, dataset means  \cite{brandon_carter_b02b21dd}, noise, and blurred inputs  \cite{ruth_c__fong_7e902699} can break temporal continuity when applied to individual frames. Such substitutions may freeze or distort the motion patterns on which video understanding models depend  \cite{avinab_saha_e5427afc}.

Our method therefore uses hot-deck-style temporal forward-fill imputation  \cite{lorenzo_beretta_292544f4}. When a frame is masked, we replace it with the closest retained frame to its left. If no retained frame exists to the left, for example, when the masked frame precedes the first retained frame, we instead use the closest retained frame to its right. Given a binary selection mask $m \in \{0,1\}^{T}$, each masked frame $I_t$ is replaced by the closest frame retained from the past: $I_{t^*}$ where $ t^* = \max\{t' < t \;:\; m_{t'} = 1\}$ i.e., temporal forward-fill. This procedure draws on an observed frame from the same video, thereby preserving the input’s temporal structure while limiting distributional shift. Although naive frame duplication has been reported to substantially alter prediction values in video-classification models  \cite{de_an_huang_7d745de8}, our approach does not duplicate frames indiscriminately. Instead, it selectively removes frames whose absence has the smallest effect on model behavior.

More sophisticated methods can generate missing frames, including ODE-based approaches  \cite{sunghyun_park_04f5615a, david_kanaa_5e16e56e}, diffusion-based methods  \cite{tobias_h_ppe_7d4f2b89, duolikun_danier_79453eb0, nicolas_cherel_7b42d3b1}, and inpainting-based methods  \cite{minhyeok_lee_1adeaba5, chaohao_xie_ec50db0c}. We do not use these methods because their generative capabilities may reconstruct information that has been intentionally removed, thereby undermining the interpretation of the resulting subsets. Our method explicitly uses the information in the retained frames to construct a new video with the original length. Moreover, many of these approaches are computationally expensive. In the section \ref{sec:comp_imp}, we empirically demonstrate that our frame-imputation method produces the smallest change in model behavior compared with other common methods.

\section{Experiments}

The four images shown in Figure \ref{fig:example} are representative examples of detected sufficient frame subsets, with no more than three subsets shown per video for clarity. In \textit{Archery}, the selected frames capture the visual evidence used by the MC3-18 model to recognize the action. In \textit{Approaching something with your camera}, two frames are sufficient: the first shows the object at a distance, while the second shows it closer, conveying the approaching motion. In \textit{Bending something until it breaks}, two frames capture the transition from an intact match to a broken one, allowing the V-JEPA 2 model to infer the action. Finally, the \textit{Digging something out of something} examples show a hand clutching the soil and the object being unearthed, highlighting the key stages supporting the classification.

\subsection{Datasets and models}
We evaluate on two complementary video understanding benchmarks: The test set of UCF101 \cite{khurram_soomro_a3c7bbc0} and Something-Something v2 \cite{raghav_goyal_0d4c7f22}. We selected six random samples from each class of the test set of the SSV2 dataset to keep the computational load manageable. We incorporate SSV2 because its classes carry markedly more temporal nuance. Whereas single-frame cues can often suffice on UCF101, SSV2 forces both the model and our frame subsets to rely on genuinely temporal evidence. We use two models per dataset each pretrained on the respective datasets, spanning convolutional and attention-based architectures. On UCF101 we use r3d-18, which applies 3D convolutions throughout the spatiotemporal volume, and mc3-18, its mixed-convolution variant combining 2D and 3D convolutions, chosen as lightweight, economical backbones  \cite{kensho_hara_bb74ffbe}. On SSV2 we use V-JEPA 2  \cite{mido_assran_c74ff668}, and TimeSformer-base, a convolution-free transformer that models spatiotemporal structure through divided space-time attention and achieves state-of-the-art results on action recognition benchmarks  \cite{gedas_bertasius_9606344a}. Together, these models contrast classic 3D convolutional temporal modeling with modern attention-based temporal reasoning. In all our experiments, model weights were frozen and the models were used in inference mode.

\subsection{Quality of frame ordering}
\label{sec:ordering_qaulity}

\begin{table*}
	\centering
	\begin{tabular}{@{}cccccccccc@{}}
		\toprule
		dataset & model & SFS & random & facility & facility-DINO & GradCAM & IG & Shapley & STAA \\ 
		\midrule
		UCF101 & mc3-18 & 0.764108 & 0.656703 & 0.65345 & 0.658763 & 0.737171 & 0.708959 & 0.657089 & ~ \\ 
		UCF101 & r3d-18 & 0.699366 & 0.556029 & 0.54905 & 0.560915 & 0.542509 & 0.596897 & 0.555181 & ~ \\ 
		SSV2 & vjeap2 & 0.782793 & 0.571896 & 0.528419 & 0.604764 & 0.510894 & 0.590939 & 0.573097 & 0.467519 \\ 
		SSV2 & tf-b & 0.656744 & 0.45991 & 0.472702 & 0.498369 & 0.409654 & 0.477168 & 0.462727 & 0.384753 \\ 
		\bottomrule
	\end{tabular}
	\caption{Insertion AUC of output probability of the predicted class}
	\label{tab:insertion}
\end{table*}

\begin{table*}
	\centering
	\begin{tabular}{@{}cccccccccc@{}}
		\toprule
		dataset & model & SFS & gradcam & IG & random & Shapley & facility-DINO & facility & STAA \\
		\midrule
		UCF101 & mc3-18 & -0.4060 & -0.2901 & -0.1656 & 0.1276 & 0.1128 & 0.0959 & 0.1610 & ~ \\
		UCF101 & r3d-18 & -0.2603 & 0.2460 & 0.0998 & 0.2108 & 0.2218 & 0.1986 & 0.2427 & ~ \\
		SSV2 & vjepa2 & -0.0987 & 0.6445 & 0.5038 & 0.5476 & 0.5452 & 0.4630 & 0.7539 & 0.7467 \\
		SSV2 & tf-b & -0.0987 & 0.6445 & 0.5038 & 0.5476 & 0.5452 & 0.4630 & 0.7539 & 0.7467 \\
		\bottomrule
	\end{tabular}
	\caption{Sufficiency metric}
	\label{tab:suff}
\end{table*}

Our SFS algorithm first ranks frames by importance and then uses this ranking to extract the first SFS. Accurate frame ordering is essential for producing concise and interpretable subsets. We evaluate each ordering by measuring how quickly the model’s softmax probability for the predicted class increases as frames are added sequentially. An effective ordering places the most informative frames first, resulting in a steeper initial increase and a larger area under the probability curve.

The table \ref{tab:insertion} compares insertion AUC across the following approaches. \textbf{SFS} denotes our sufficient-frame-subset method. \textbf{Random} randomly orders the frames as a baseline. \textbf{Facility} applies the facility-location algorithm  \cite{david_b__shmoys_52bcbcfa} to cluster feature embeddings and order frames by their representativeness. This approach has also been used to identify meaningful input tokens for video understanding in an state-of-the-art manner  \cite{janghoon_cho_62c5bc88}. We extract embeddings for facility location in two ways. First, each frame is repeated to construct a complete video, which is then passed through the corresponding video model. Because this procedure may introduce model-dependent bias, we also evaluate \textbf{Facility-DINO}, which applies facility location to embeddings extracted from DINOv2  \cite{maxime_oquab_f23ff8f7} rather than from the task-specific video model. \textbf{Grad-CAM} orders frames according to their class-specific, gradient-based importance, following a widely used approach for visualizing model decisions  \cite{ramprasaath_r__selvaraju_bb3d7670}. \textbf{IG} denotes Integrated Gradients, an axiomatic attribution method  \cite{mukund_sundararajan_62086538}. Shapley estimates each frame’s contribution to the prediction using a sampling-based approximation of the Shapley value  \cite{strumbeljerik_42cc2316, javier_castro_28f89a07, lloyd_s__shapley_005c222c, hugh_chen_5eacd5cb}. This approach is the closest to the method introduced in Play Fair  \cite{will_price_1d4ff25d}. However, Play Fair is not directly applicable to our setting because it relies on a multi-scale model, whereas our model-agnostic method does not require such an architecture. \textbf{STAA} is a state-of-the-art transformer-specific attribution method that derives feature importance from attention weights. We aggregate attention across heads and spatial dimensions to obtain frame-level attention scores  \cite{zerui_wang_6183c62e}. STAA is applicable only to the two transformer-based models evaluated on the SSV2 dataset. 

The results show that SFS achieves the highest insertion AUC, indicating that it places the most decision-relevant frames earliest and produces the fastest increase in the model’s predicted-class probability. The lower AUC values obtained by the competing methods suggest that their rankings are less effective at prioritizing frames that are sufficient for the model’s decision. Facility and Facility-DINO obtain similar AUC values, suggesting that the potential bias introduced by constructing classifier inputs through frame repetition has a limited effect on the resulting frame ordering.

\subsection{Quality of the SFS}
We calculate the sufficiency metric  \cite{jay_deyoung_a7ccba42} to assess how well the selected frame subsets preserve the model’s decision. Specifically, we measure the relative drop in the predicted-class probability as $sufficiency = \frac{\Phi(V)_c - \Phi(V_S)_c}{\Phi(V)_c}$. Lower values indicate higher-quality frame subsets because they preserve more of the original model’s predictive behavior. We use the first sufficient frame subset identified by SFS and compare it with a subset containing the same number of top-ranked frames selected by each competing method. Across all experimental settings, SFS produces the highest-quality frame subsets (Table \ref{tab:suff}). Note that negative values are allowed for sufficiency according to our definition because it indicates that the chosen frame subset actually improved the classification probability. 

\subsection{Effect of threshold on the brevity of explanations and overall accuracy}
Table \ref{tab:thr} examine the trade-off between predictive accuracy and explanation brevity across different sufficiency thresholds. Table \ref{tab:thr} reports the classification accuracy achieved using sufficient frame subsets extracted at each threshold $\tau$. Lower thresholds impose stricter sufficiency requirements and therefore retain more of the original model’s predictive information. By contrast, higher thresholds permit greater deviations from the full-video prediction, resulting in a gradual decline in accuracy. Table \ref{tab:thr} reports the average number of frames in a SFS at each threshold and shows that higher thresholds generally produce shorter explanations. Together, these results demonstrate the trade-off between preserving predictive performance and minimizing the number of frames used for explanation, thereby informing the selection of a dataset-level threshold. At the individual-video level, these statistics can also support the use of a locally adapted threshold according to the desired balance between accuracy and brevity of the explanation. Because the choice of threshold is inherently subjective, we use a threshold of 0.1 in all subsequent experiments, as it provides a favorable balance between accuracy retention and interpretability.

\begin{table*}
	\centering
	\setlength{\tabcolsep}{4pt}
	\begin{tabular}{@{}ccccccccc@{}}
		\toprule
		dataset & model & 0 & 0.1 & 0.2 & 0.3 & 0.4 & 0.5 & original \\
		\midrule
		UCF101 & mc3-18 & 0.84/1.2 & 0.84/1.1 & 0.84/1.1 & 0.84/1.0 & 0.84/1.0 & 0.84/1.0 & 0.85/16 \\
		UCF101 & r3d-18 & 0.81/4.8 & 0.81/3.4 & 0.81/3.1 & 0.81/2.9 & 0.80/2.7 & 0.80/2.5 & 0.81/16 \\
		SSV2 & vjepa2 & 0.67/4.8 & 0.68/3.4 & 0.68/3.1 & 0.68/2.9 & 0.68/2.7 & 0.67/2.5 & 0.67/16 \\
		SSV2 & tf-b & 0.48/3.4 & 0.48/2.8 & 0.48/2.7 & 0.47/2.5 & 0.47/2.4 & 0.46/2.3 & 0.48/8 \\
		\bottomrule
	\end{tabular}
	\caption{Accuracy/num frames in a subset under different thresholds ($\tau$). Original column shows model accuracy with all the frames and the number of frames used for a full video}
	\label{tab:thr}
\end{table*}

\subsection{Comparing various methods for frame imputation}
\label{sec:comp_imp}
\begin{table}
	\centering
	\begin{tabular}{@{}ccccc@{}}
		\toprule
		       & \multicolumn{2}{c}{UCF101} & \multicolumn{2}{c}{SSv2}\\
		Method & JS & MMD & JS & MMD \\ 
		\midrule
		future & 3.7 & 0.04 & 116.3 & 14.2 \\ 
		past & 3.6 & 0.04 & 107.6 & 12.3 \\ 
		zero & 10.6 & 5.09 & 189.7 & 19.9 \\ 
		mean & 10.9 & 5.59 & 187.4 & 20.5 \\ 
		interp & 4.2 & 0.13 & 143.6 & 20.6 \\ 
		\bottomrule
	\end{tabular}
	\caption{Distribution change cause by Various frame imputation methods. Shown values are multiplied by $10^4$}
\end{table}

We examined how replacing a missing frame with different common imputation methods affects model behavior. We measured this effect using two metrics. \textbf{JS} denotes the Jensen–Shannon divergence  \cite{jinfeng_lin_5004d5e1} between the original and imputed classification-probability distributions. \textbf{MMD}, or Maximum Mean Discrepancy  \cite{krikamol_muandet_69c87e11}, measures the distance between the embedding distributions of the original and imputed videos. Lower values for both metrics indicate greater similarity between the original and imputed distributions. 

We use the MC3-18 model on the UCF101 dataset and the V-JEPA 2 model on the SSV2 dataset to obtain embeddings and classification predictions. We evaluate the following imputation methods: \textbf{Past}, which uses the procedure described in the Method section; \textbf{Future}, which uses frames from the future portion of the video to impute missing frames; \textbf{Zero}, which replaces missing frames with zeros; \textbf{Mean}, which replaces them with the mean values of the available frames; and \textbf{Interp}, which temporally interpolates the missing pixel values from the corresponding pixels in neighboring frames. The results show that the \textbf{Future} and \textbf{Past} methods produce imputed videos that are substantially more similar to the original videos than the zero, mean, and interpolation baselines, as indicated by their lower JS and MMD values. These findings suggest that temporally adjacent frames preserve the model’s predictive behavior and embedding structure more effectively than generic replacement values or pixel-wise interpolation.

\subsection{Cross-model transferability of the Frame Subsets}

\begin{figure}
	\includegraphics[width=0.9\columnwidth]{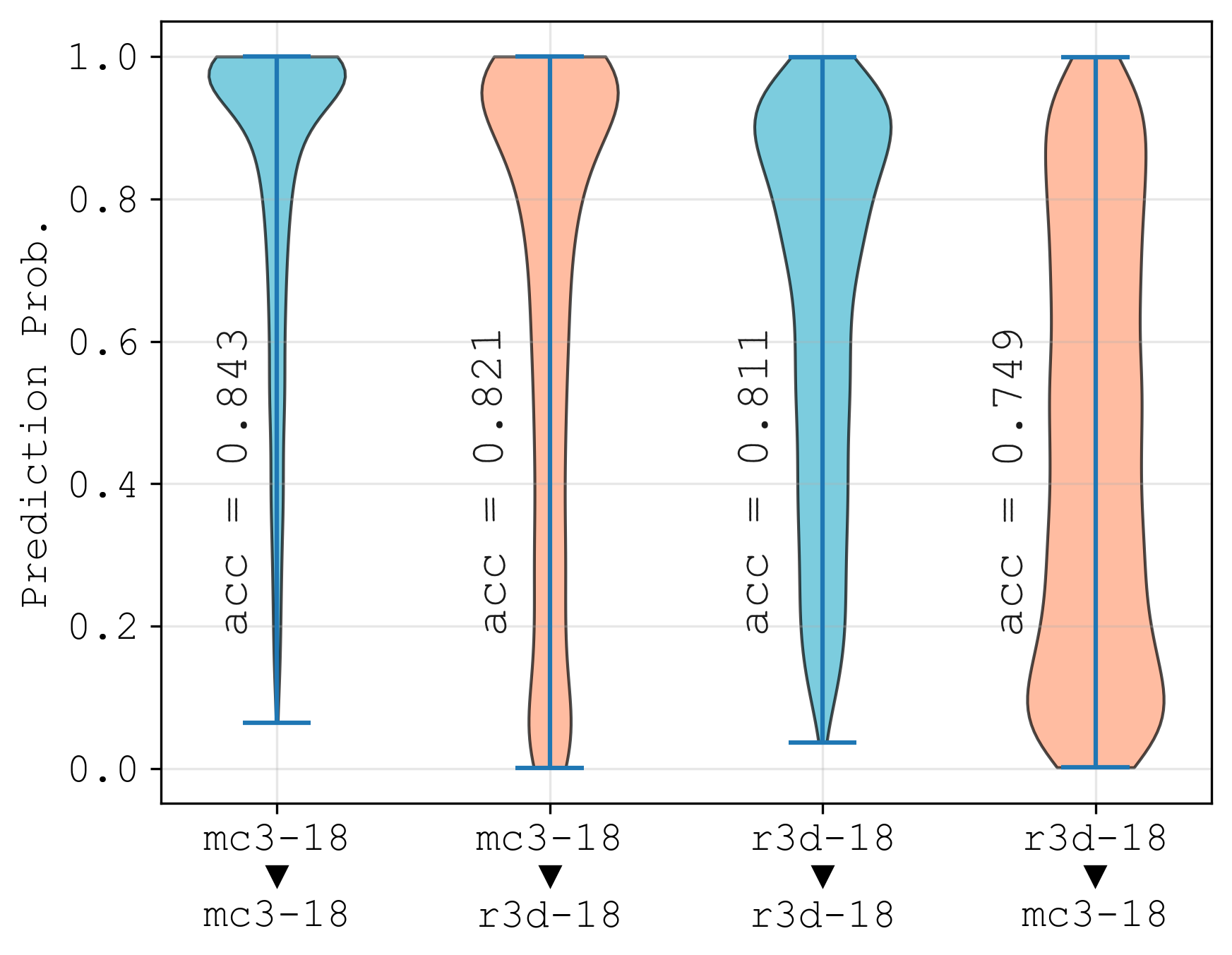}
	\caption{Using subsets detected with one model on another for UCF101 dataset}
	\vspace{-15pt}
	\label{tab:transfer}
\end{figure}

The figure \ref{tab:transfer} examines whether sufficient frame subsets are interchangeable across models on UCF101. Classification accuracy declines substantially when a frame subset identified by one model is evaluated using another, indicating limited cross-model transferability. Frame subsets identified by MC3-18 appear more robust and higher quality: when applied to R3D-18, they produce only a modest accuracy decrease, whereas applying R3D-18-selected frame subsets to MC3-18 results in a considerably larger decline. This asymmetry may be related to MC3-18’s superior predictive accuracy, which could enable it to identify more reliable and transferable sufficient frame subsets. However, differences in the models’ architectural biases or training objectives may also contribute to this effect. Disentangling these factors through further empirical investigation is an important direction for future work.

\begin{figure*}	
	\centering
	\begin{minipage}{0.35\textwidth}
		\centering
		\includegraphics[width=\linewidth]{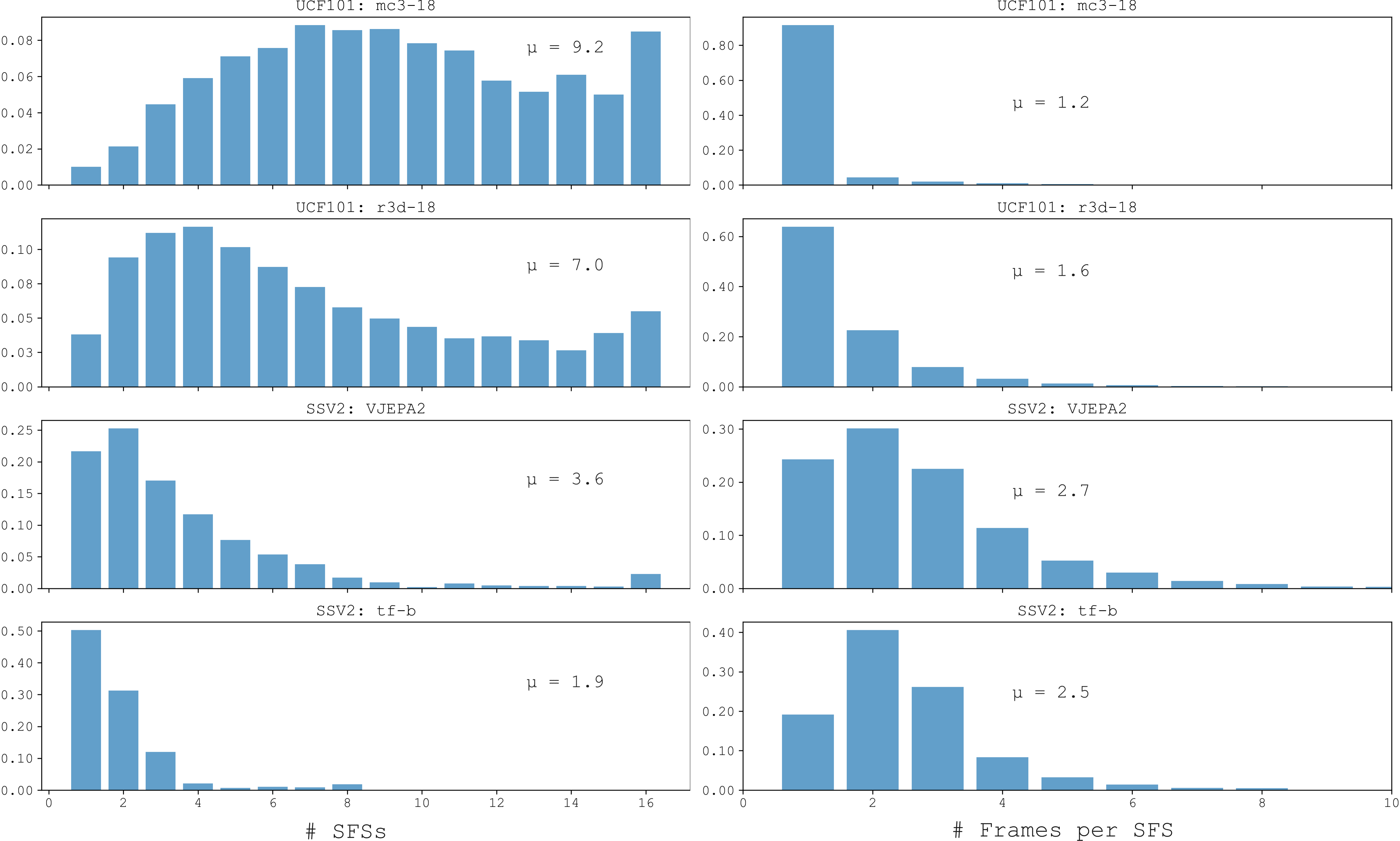}
		\caption{Frame subset statistics}
		\label{tab:stat}
	\end{minipage}
	\hfill
	\begin{minipage}{0.63\textwidth}
		\centering
		\includegraphics[width=\linewidth]{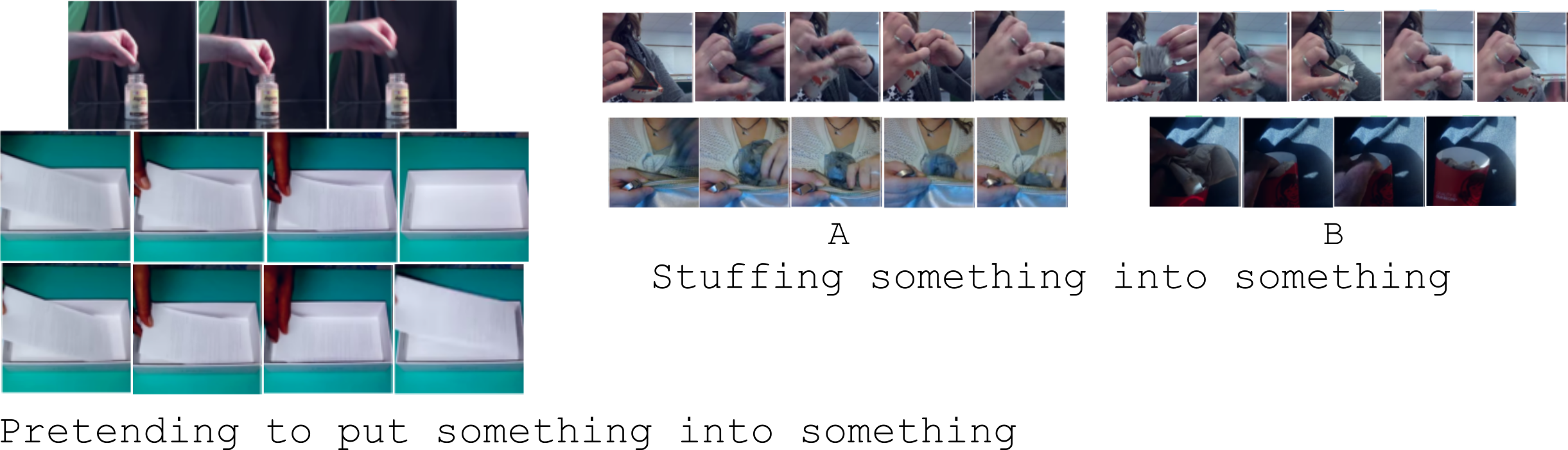}
		\caption{Several Example frame subsets}
		\label{tab:ssv2_failure}
	\end{minipage}
	
\end{figure*}

The figure \ref{tab:stat} shows that the number and average length of sufficient frame subsets vary across dataset–model combinations. UCF101 generally requires fewer frames per subset, suggesting greater reliance on static visual evidence, whereas SSV2 requires longer subsets, reflecting its stronger temporal demands. Across models, higher average numbers of sufficient frame subsets are associated with better accuracy; for example, V-JEPA 2 achieves 0.67 accuracy with 3.6 subsets on average, compared with 0.48 accuracy and 1.9 average subsets for tf-b. This pattern is consistent with findings from prior work  \cite{keisuke_kawano_e36177e0} on minimal sufficient views, which reports that models relying on a larger number of sufficient evidence subsets tend to exhibit better generalization performance. Thus, the observed relationship may reflect a fundamental property of model reasoning rather than a dataset-specific artifact. Using evidence from multiple temporal segments may support more robust predictions.

\subsection{Failure Analysis via Frame Subset}

\begin{figure}
	\centering
	\includegraphics[width=\columnwidth]{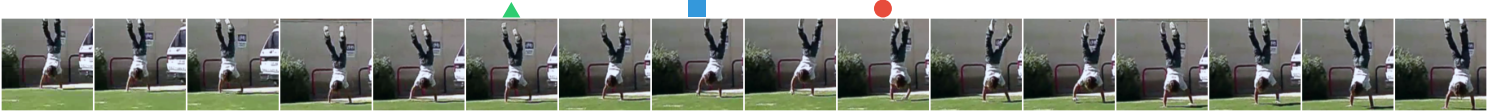}
	\includegraphics[width=\columnwidth]{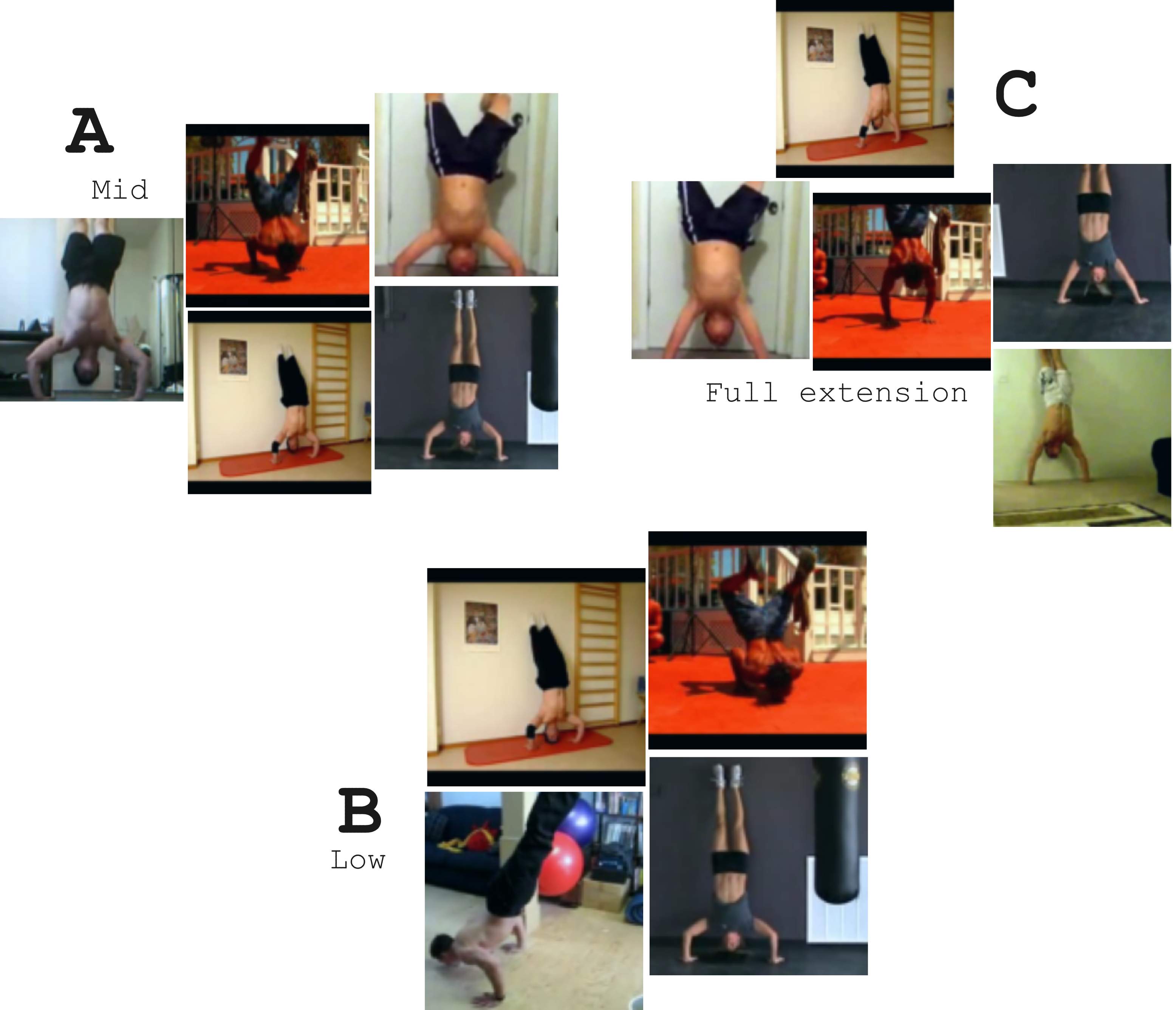}
	\caption{top: handstand walking incorrectly classified as handstand pushup. Bottom: some common SFSs for handstand pushups}
	\vspace{-12pt}
	\label{fig:ucf_failure}
\end{figure}

This figure \ref{fig:ucf_failure} presents an example from the UCF101 dataset in which the ground-truth class, \textit{handstand walking}, was incorrectly classified as \textit{handstand pushup} by the MC3-18 model. The \textit{handstand walking} class had a low average accuracy of only 20\%, and nearly 30\% of its misclassified samples were labeled as \textit{handstand pushup}. Most sufficient frame subsets in UCF101 contained only a single frame. We manually clustered several SFSs detected for \textit{handstand pushup}, identifying three broad groups based on the degree of arm extension. The SFSs detected for \textit{handstand walking} could easily be mistaken for members of these clusters. This suggests that the model may not capture enough temporal information to reliably distinguish \textit{handstand walking} from \textit{handstand pushup}.

\begin{figure*}
	\centering
	\includegraphics[width=0.9\textwidth]{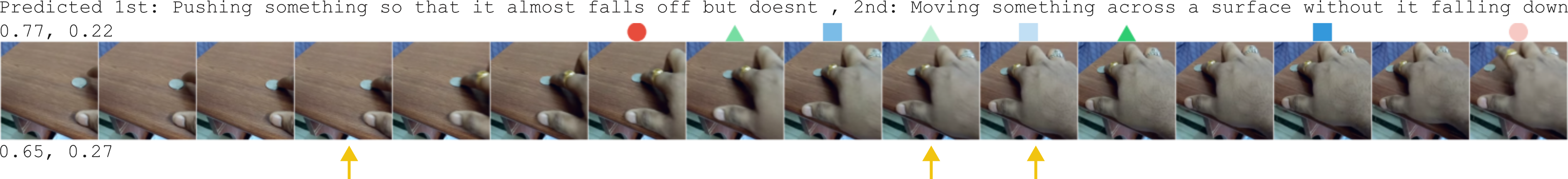}
	\caption{SSV2 incorrect classification: GT: moving sth across a surface without it falling down. Pred: pushing sth so that it almost falls off but doesn't. Shapes above show the SFSs for the predicted class. Arrows at the bottom shows the SFSs calculated for the GT class. Prediction probability of the GT class improves from 0.22 to 0.27 after taking the SFSs for the GT class}
	\label{tab:pull}
	
	\vspace{1pt}
	
	\centering
	\includegraphics[width=0.9\textwidth]{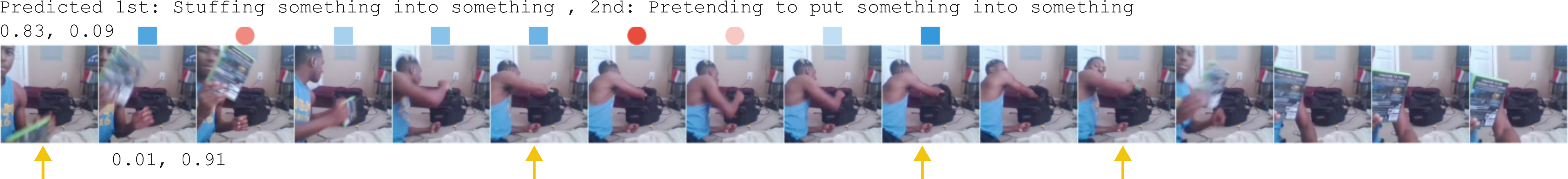}
	\caption{SSV2 incorrect classification: GT:pretending to put sth into something, pred: stuffing sth into sth, Arrows at the bottom shows the SFSs calculated for the GT class. Prediction probability of the GT class improves from 0.09 to 0.91 after taking the SFSs for the GT class}
	\label{tab:pretend}

\end{figure*}

The figure \ref{tab:pull} shows a sample from the SSV2 dataset with the ground-truth label c1, “moving something across a surface without it falling down.” The V-JEPA 2 model incorrectly classified it as c2, “pushing something so that it almost falls off but doesn’t.” Three sufficient frame subsets for c2 are shown above, while the arrows at the bottom indicate a SFS supporting class c1. The earlier frames of the video are consistent with “moving something across a surface without it falling down,” whereas the later frames provide evidence for “pushing something so that it almost falls off but doesn’t,” as the coin is pushed to the very edge of the table. The c2 SFS primarily captures the latter part of the video, leading the model to predict c2. In contrast, the SFS identified for c1, shown with the yellow arrows, focuses on frames in which the coin is pushed across the surface and largely excludes the later frames near the edge.

The figure \ref{tab:pretend} shows a sample from the SSV2 dataset with the ground-truth label c1, “pretending to put something into something.” The V-JEPA 2 model incorrectly classifies it as c2, “stuffing something into something.” The SFSs identified for c2 focus mainly on the early part of the video. Consequently, the model overlooks the person removing the object from the backpack, leading it to interpret the action as genuine rather than pretend. In contrast, the SFS identified for c1 includes frame 11, which shows the person taking the object back. This subset captures the evidence needed to correctly classify the video as c1.

We manually clustered several frame subsets generated by the V-JEPA 2 model for classes c1 and c2. For class c1, the SFSs capture three main stages of the action: approaching the object and positioning it where it will be placed, putting it in or on the target location, and subsequently taking it back. For class c2, the SFSs form two primary clusters. Cluster B includes later frames to determine whether the object remains in the container or whether the person takes it back. In contrast, Cluster A excludes the later frames and therefore does not account for whether the object is retrieved. The model appears to have learned to associate patterns such as those in Cluster A with class c2. Consequently, it may incorrectly classify videos in which the person takes the object back, because the model does not consider the later frames that provide this disambiguating evidence.

\FloatBarrier

\subsection{Limitations}
Some of the identified SFSs may provide redundant explanations, but our current formulation cannot detect or remove this redundancy. The SFS algorithm is also computationally expensive: analyzing a 16-frame video requires 135 model evaluations. This cost could potentially be reduced by exploiting the high redundancy among adjacent video frames. For example, if two frames are highly correlated, the model may respond similarly to both. Developing methods to identify and eliminate redundant frames is an important direction for future research.

\clearpage
{
    \small
    \bibliographystyle{ieeenat_fullname}
    \bibliography{main}
}

\end{document}